\documentclass{article}

\usepackage{microtype}
\usepackage{graphicx}
\usepackage{subfigure}
\usepackage{booktabs} 
\usepackage{amsmath}
\usepackage{amsfonts}
\usepackage{array}
\usepackage[ruled,vlined,algo2e]{algorithm2e}
\usepackage{xcolor}

\usepackage{hyperref}

\newcommand{\cut}[1]{}
\newcommand{\modelname}{Feature-Critic}

\newcommand{\keypoint}[1]{\vspace{0.0cm}\noindent\textbf{#1}\quad}

\newcommand{\nice}[1]{\textcolor{black}{#1}}

\usepackage{multirow}

\usepackage[accepted]{icml2019}

\icmltitlerunning{Feature-Critic Networks for Heterogeneous Domain Generalisation}

\begin{document}

\twocolumn[
\icmltitle{Feature-Critic Networks for Heterogeneous Domain Generalisation}



\icmlsetsymbol{equal}{*}

\begin{icmlauthorlist}
\icmlauthor{Yiying Li}{equal,nudt,ed}
\icmlauthor{Yongxin Yang}{equal,ed}
\icmlauthor{Wei Zhou}{nudt}
\icmlauthor{Timothy M. Hospedales}{ed,saic}
\end{icmlauthorlist}

\icmlaffiliation{nudt}{College of Computer, National University of Defense Technology, Hunan, China}
\icmlaffiliation{ed}{School of Informatics, The University of Edinburgh, Edinburgh, UK}
\icmlaffiliation{saic}{Samsung AI Centre, Cambridge, UK}

\icmlcorrespondingauthor{Timothy M. Hospedales}{t.hospedales@ed.ac.uk}

\icmlkeywords{Machine Learning, ICML}

\vskip 0.3in
]



\printAffiliationsAndNotice{\icmlEqualContribution} 

\begin{abstract}
The well known domain shift issue causes model performance to degrade when deployed to a new target domain with different statistics to training. Domain adaptation techniques alleviate this, but need some instances from the target domain to drive adaptation. Domain generalisation is the recently topical problem of learning a model that generalises to unseen domains out of the box, and various approaches aim to train a domain-invariant feature extractor, typically by adding some manually designed losses. In this work, we propose a \emph{learning to learn} approach, where the auxiliary loss that helps generalisation is itself learned. Beyond conventional domain generalisation, we consider a more challenging setting of \emph{heterogeneous} domain generalisation, where the unseen domains do not share label space with the seen ones, and the goal is to train a feature representation that is useful off-the-shelf for novel data and novel categories. Experimental evaluation demonstrates that our method outperforms state-of-the-art solutions in both settings.
\end{abstract}

\section{Introduction}

A shift in data statistics between training and testing is often unavoidable in real-world applications, and leads to a significant negative impact on the performance of machine learning models in practice. This motivates research into methods to ameliorate the impact of domain shift, including  Domain Adaption (DA) \cite{Bousmalis16, Ganin15, Long15, Long16} and Domain Generalisation (DG) \cite{Muandet13, Ghifary15, Li18, Shankar18}.

Unsupervised Domain Adaptation (UDA) \cite{Long16, Saito17tri, Shu18dirt} methods operate in the setting where we can access unlabelled testing (target) domain data during training to drive model adaptation and compensate for the domain shift. Domain Generalisation addresses the harder setting, where a model trained on a set of source domains should perform well on a novel target domain with different data statistics, without requiring any access to target domain data during training. That is, the model should be robust enough  out-of-the-box to perform well in a new domain, without further parameter updates. Both DA and DG methods almost always assume the label space is consistent across both source and target domains.

In the case of disjoint label spaces between source and target domain, we term the domain generalisation problem as one of \emph{heterogeneous} domain generalisation. In this case a \emph{feature representation} trained on a source domain should generalise to supporting recognition of novel categories in a novel target domain. This problem setting is actually widely encountered. The central example is the ubiquitous computer vision pipeline where a CNN feature extractor pre-trained on ImageNet is re-used for diverse applications. If data, computation, and human expert time is available, the feature can be fine-tuned on the target problem. However, for many practical applications lacking one or more of these requirements, standard practice is to use an ImageNet CNN off-the-shelf as a  fixed feature extractor, and  train a shallow model such as SVM or KNN for the new problem  \cite{donahue2014decaf,razavian2014cnnAstound}. This pipeline is an example of the heterogeneous domain generalisation setting, in that a feature is being asked to generalise to supporting recognition of novel categories in data with novel statistics. The ImageNet pre-trained feature is strong enough to do a reasonable job of this already. However, given the ubiquity of this pipeline, providing an improved general purpose feature would be widely beneficial. In this paper we aim to do exactly this by presenting a novel method that explicitly  trains a feature to prepare it for domain and label shift. We demonstrate this via performing heterogeneous DG on the Visual Decathalon benchmark \cite{Rebuffi17}. This also provides the largest scale evaluation of DG to date.

We are inspired by recent meta-learning learning methods that perform episodic training \cite{Finn17, snell2017prototypicalNets,ravi2017fewShotMeta} to simulate the train/test process to improve few-shot learning. In this work, we propose to perform meta-learning to improve feature extractor training, and deliver a better model for both homogeneous and heterogeneous DG problems. 

To realise our idea, we simulate training-to-testing domain shift  by splitting our source domains into virtual training and testing (i.e., validation) domains. The source model is decomposed into feature extractor  and task networks (i.e., a classifier network in our case). Crucially we then introduce a feature-critic network that learns to criticise the quality of the features produced by the feature network, specifically with regards to their robustness to the simulated domain shift. This feature-critic provides a learned auxiliary loss which provides an additional source of feedback to the feature network (besides the conventional supervised classification loss via the task network), and enables it to produce a more robust feature. The feature, task and critic networks are trained together end-to-end in a meta-learning pipeline. Our evaluation shows good performance in the conventional DG setting using  Rotated MNIST \cite{Ghifary15, Motiian17} and PACS \cite{Lida17} benchmarks, as well as the heterogeneous DG setting using the larger scale  Visual Decathlon (VD) \cite{Rebuffi17} benchmark.

\section{Related Work}

\keypoint{Multi-Domain Learning (MDL)} MDL addresses training a single model capable of solving multiple datasets (domains). If the data is relatively small and the domains are similar, this sharing can lead to improved performance compared to training a separate model per domain \cite{Yang15}. On the other hand, for diverse domains with large data, MDL may under-perform a single model per domain; but is nonetheless is of interest due to the simplicity of a single model and its better memory scalability compared to a separate model per domain \cite{Rebuffi17, Rebuffi18}. We mention MDL here, because  DG methods typically train on multiple source domains as per MDL -- but furthermore aim to generalise to novel held out domains.

\keypoint{Domain Generalisation (DG)} DG relates to domain-adaptation in that we care about performance on a target domain, rather than source domains; however it considers the case where target domain samples are unavailable during training, so the model must generalise directly rather than adapt to the target domain. DG is of related to conventional generalisation: where models learned on a set of training instances  generalise to novel testing instances, for example by regularisation. \nice{However it operates at a higher level, where we aim to help models trained on a set of training domains generalise to a novel testing domain.} 

Most existing DG approaches can be split into three categories: feature-based methods, classifier-based methods, and data augmentation methods. Feature-based methods: These aim to generate a domain-invariant  representation. For example where the distance between the empirical distributions of the source and target examples is minimized \cite{LiH18, Muandet13,Li18}. Classifier-based methods: Some aim to enhance generalisation by fusing multiple sub-classifiers learned from source domains \cite{Duan12, Niu15, NiuL15}, and others learn an improved classifier regulariser using source samples -- notably the recently proposed MetaReg \cite{Balaji18}. Data augmentation methods: CrossGrad \cite{Shankar18} generates provides domain-guided perturbations of input instances, which are then used to train a more robust model. \citet{Volpi18} defines an adaptive data augmentation scheme by appending adversarial examples at each iteration. Our \modelname{} approach falls into the feature-based category, but meta-learns a feature-critic network to train a robust shared feature extractor.

Few studies have considered the heterogeneous DG setting, where the domains do not share the same label space. We do not expect the \emph{classifier} to generalise directly to the target domain (impossible due to the change in label space), but we do aim to improve the robustness of a source-domain trained \emph{feature} in terms of its generalisation to successfully represent a novel problem. Most existing DG methods cannot be applied here\cut{ besides Domain Adaptive Neural Networks \cite{Ganin16} and CrossGrad \cite{Shankar18}}. We show how to modify MetaReg \cite{Balaji18} and Reptile \cite{nichol2018reptileFOML} to address this DG setting. The most relevant benchmark is Visual Decathlon (VD) \cite{Rebuffi17}. The VD benchmark was proposed to evaluate multi-domain and lifelong \cite{Rosenfeld18} learning. \cut{VD competitors should originally learn a model covering all ten domains, with low parameter growth. } We re-purpose it for DG evaluation. In this case a model trained on the six largest datasets in VD should produce a feature which provides a general and robust enough encoding to allow the four smaller datasets to be classified with a simple shallow classifier.

\keypoint{Meta-Learning} Meta-learning (a.k.a. learning to learn, \cite{schmidhuber1997inductiveBias,thrun1998learntolearn}) has received resurgence in interest recently with applications in few-shot  learning \cite{Li17,snell2017prototypicalNets,Sung18} and beyond \cite{Xu18}. In few-shot meta-learning, a common strategy is to simulate the few-shot learning scenario by randomly drawing few-shot train/test episodes from the full training set. We adapt this episodic training strategy by creating virtual training and testing splits of our source domains in each mini-batch. 

A few methods have applied related episodic meta-learning strategies in DG \cite{Li18,Balaji18}. MLDG \cite{Li18} defined a heuristic gradient descent update rule based on the gradients of the simulated training and testing domains. MetaReg \cite{Balaji18} trains the weights of the \emph{classifier}'s regulariser so as to produce a more general classifier for a fixed feature extractor. In contrast,  our \modelname{} produces a more general \emph{feature extractor} that can be used with any classifier. This is achieved by simultaneously learning an auxiliary loss function \cite{gygli2017deepValueNet,yang2017metacritic} (i.e., the critic network) that trains the feature extractor for improved domain invariance.

\begin{figure}[t]
\centering
\includegraphics[width=0.48\textwidth]{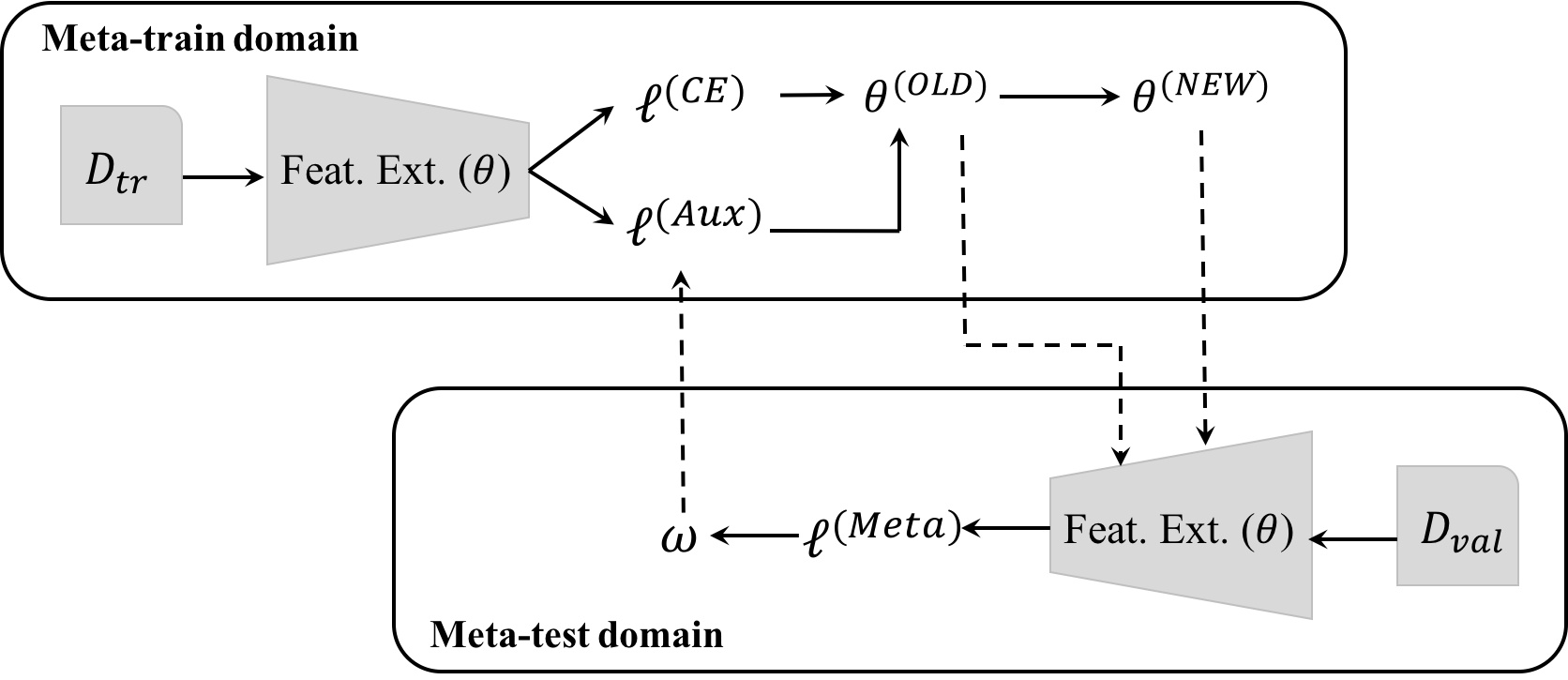}
\caption{Illustration of the \modelname{} learning framework.} \label{fig:illu}
\vspace{-1.0em}
\end{figure}

\section{Methodology}

We  introduce the proposed method under the heterogeneous DG setting, but it is straightforwardly applicable to conventional (homogeneous) DG as a special case. Assuming that we have N domains (datasets) $\mathcal{D}=\{D_1, D_2, \dots, D_N\}$, and each domain contains a set of data-label pairs, i.e., $D_i=\{X^{(i)}, Y^{(i)}\}$. We also have the training split of target (testing) domain, $D_{N+1}=\{X^{(N+1)}, Y^{(N+1)}\}$, but  we can not access this for feature learning. 

We assume a CNN model split into two parts: feature extractor $f_\theta$ and classifier $g_\phi$. For heterogeneous DG, we have $N$ classifiers, denoted $g_{\phi_1}, g_{\phi_2}, \dots, g_{\phi_N}$, and a universal feature extractor $f_\theta$  shared for all domains (assuming that images from all domains are resized to the same size). In the homogeneous DG, we only need a single classifier $g_\phi$ that can be shared across all domains.

The proposed workflow is: (i) train a multi-domain model $g\circ f$ with $\mathcal{D}$, (ii) take the shared feature extractor part $f_\theta$ and use it as a fixed feature extractor for the target domain, (iii) extract features for target domain's train set $D_{N+1}$ and train a SVM or KNN classifier, (iv) evaluate on the testing split of the target domain, denoted as $\tilde{D}_{N+1}=\{\tilde{X}^{(N+1)}, \tilde{Y}^{(N+1)}\}$. In the case of homogeneous DG, we can also take the shared classifier $f_\phi$ and use the full model $g_\phi \circ f_\theta$ directly for the target domain. The goal is to perform the training in step (i) above so that the feature extractor $f_\theta$ is robust enough to perform well on any target domain without fine-tuning.

\subsection{A Simple Baseline}

A naive deep learning approach called aggregation (AGG) trains a single extractor to minimise the total  cross-entropy (CE) loss of all domains. 
\begin{equation}\label{eq:basic}
\underset{\theta,\phi_1,\phi_2,\dots,\phi_N}{\operatorname{min}}~~~ \sum_{D_j\in \mathcal{D}}\sum_{d_j\in D_j}\ell^{(\text{CE})}(g_{\phi_j}(f_\theta(x^{(j)})), y^{(j)})
\end{equation}
Here $D_j$ is the $j$th domain and $d_j$ is a mini-batch of it. 

Then we fix $f_\theta$ and extract features for the training split of target domain $D_{N+1}$, i.e., $f_\theta(X^{(N+1)})$. With those extracted features, we can train a classifier using the pairs $\{f_\theta(X^{(N+1)}), Y^{(N+1)}\}$. Finally we test the model on the testing split of target domain $\tilde{D}_{N+1}=\{\tilde{X}^{(N+1)}, \tilde{Y}^{(N+1)}\}$.

This simple baseline surpasses many prior purpose designed DG methods as discussed in \citet{Lida17}. The key question is how to improve this naive approach, such that the trained feature extractor $f_\theta$  produces more robust features that generalise better to unseen target domains.

\subsection{Simulating Domain Shift in Training}

Our high-level strategy simulates domain-shift during training  as  illustrated in Algo.~\ref{alg1}  and Figure~\ref{fig:illu}. We use the learned feature-critic loss to guide learning on the meta-training set $\mathcal{D}_\text{trn}$, and optimise the feature-critic itself on the meta-validation set $\mathcal{D}_\text{val}$. The key idea is that training $f_\theta$ with $h_\omega$ on $\mathcal{D}_\text{trn}$ should improve its performance on $\mathcal{D}_\text{val}$.

\begin{algorithm}[t]
	\DontPrintSemicolon
	\KwIn{$\mathcal{D} = \{D_1, D_2, \dots, D_N\}$}
	\KwOut{$\theta$}
	\Begin{
		\While{not converge or reach max steps}{

			\textbf{Randomly split} $\mathcal{D}$:
			
			$\mathcal{D}_\text{trn} \cap \mathcal{D}_\text{val} = \emptyset$
			
			$\mathcal{D}_\text{trn} \cup \mathcal{D}_\text{val} = \mathcal{D}$
			
			\For{$t\in [1,2,\dots,T]$}{

				Sample mini-batch $D_{tr}$\cut{$=\{x^{(j)}, y^{(j)}\}$} from each $D_j\in \mathcal{D}_\text{trn}$
				
				Optimise feature extractor $g_\theta$ on $D_{tr}$ using supervised and auxiliary loss $h_w$.

				Sample mini-batch $D_{val}$\cut{$=\{x^{(k)}, y^{(k)}\}$} from each $D_k\in \mathcal{D}_\text{val}$
				
                Optimise auxiliary loss $h_\omega$ on $D_{val}$
				
				}
		}
	}
	\caption{Simulating Domain Shift in Training}\label{alg1}
\end{algorithm}

\subsection{Meta-Learning an Auxiliary loss}

We aim to make the model training process behave well, i.e., after each  update with mini-batches from $\mathcal{D}_\text{trn}$, performance should improve for  mini-batches from $\mathcal{D}_\text{val}$. This would happen to some extent without any extra effort, but we aim to enforce it by introducing a learned auxiliary loss function, denoted $\ell^{(\text{Aux})}=h_\omega$. The only requirements for $h_\omega$ are (i) it outputs a non-negative scalar (since it is a loss) and (ii) its input depends on the  feature extractor's parameter $\theta$.

For now, we assume a suitable function $h_\omega$ (i.e., $\ell^{(\text{Aux})}$) exists and discuss how to use it. We discuss design choices for $h_\omega$ in Sec.~\ref{sec:h}. With the auxiliary loss, the objective function in Eq.~\ref{eq:basic} becomes,
\begin{equation}
\label{eq:advanced}
\underset{\theta,\phi_j s}{\operatorname{min}}~~~ \sum_{D_j\in \mathcal{D}_\text{trn}}\sum_{d_j\in D_j}\ell^{(\text{CE})}(g_{\phi_j}(f_\theta(x^{(j)})), y^{(j)}) + \ell^{(\text{Aux})}
\end{equation}

Taking the gradient of Eq.~\ref{eq:advanced} w.r.t. $\theta$ gives two terms: (i) cross-entropy loss and (ii) auxiliary loss (recall that $\ell^{(\text{Aux})}$'s input must depend on $\theta$, which means $\frac{\partial \ell^{(\text{Aux})}}{\partial \theta}$  generates non-zero values). 

Consider two alternative updates to $\theta$, with and without the help of $\ell^{(\text{Aux})}$. We have $\theta^{(\text{OLD})}=\theta - \alpha \frac{\partial \ell^{(\text{CE})}}{\partial \theta}$ and $\theta^{(\text{NEW})}=\theta - \alpha \frac{\partial \ell^{(\text{CE})}}{\partial \theta} - \alpha \frac{\partial \ell^{(\text{Aux})}}{\partial \theta}$. Here $\alpha$ is the step size of $\theta$. If the auxiliary loss $\ell^{(\text{Aux})}$ indeed does a good job of promoting domain invariance of $\theta$, then the latter update $\theta^{\text{(NEW)}}$ exploiting $\ell^{(\text{Aux})}$ should produce a more effective feature on mini-batches from the validation domain $\mathcal{D}_\text{val}$.

Thus we train the auxiliary loss (feature-critic network) to promote this. Specifically, we optimise the parameter $\omega$ of feature-critic network as follows: 
\begin{equation}\label{eq:omega}
\begin{split}
\underset{\omega}{\operatorname{max}}~~~  \sum_{D_j\in \mathcal{D}_\text{val}}\sum_{d_j\in D_j}\tanh(&\gamma(\theta^{(\text{NEW})}, \phi_j, x^{(j)}, y^{(j)})\\
-&\gamma(\theta^{(\text{OLD})}, \phi_j, x^{(j)}, y^{(j)}))    
\end{split}
\end{equation}
\noindent Here $\gamma$ is a function that measures the validation domain performance (larger is better), and we discuss how to design it in Sec.~\ref{sec:g}. $\tanh$ is a utility function, which converts the reward (performance gain) to utility. It reflects the commonly accepted idea concept \emph{diminishing marginal utility}, and links   $\theta^{(\text{NEW})}$ with $\theta^{(\text{OLD})}$. If $\tanh$ and the $\theta^{(\text{OLD})}$ term are excluded in  Eq.~\ref{eq:omega}, it would simply maximise the  validation set performance with $\theta^{(\text{NEW})}$. The reason Eq.~\ref{eq:omega} is better is that  $\gamma(\theta^{(\text{OLD})}, \phi_j, x^{(j)}, y^{(j)})$ serves as a baseline, making the value range -- and thus the gradient -- more stable. One can understand the role of  $\tanh$ here as a smoother version of min/max-margin or a softer version of gradient clipping.

In summary, optimising the feature-critic $h_\omega$ as Eq.~\ref{eq:omega} produces a loss $\ell^{(\text{Aux})}$ that encourages  the base network to extract domain agnostic features when applied in Eq.~\ref{eq:advanced}.

\subsection{Measuring Validation Performance: Designing $\gamma$}
\label{sec:g}
To measure validation performance, $\gamma$ can take up to four variables as input: feature extractor parameter $\theta$, classifier parameter $\phi$, data $x$, and label $y$, i.e., $\gamma(\theta,\phi,x,y)$. One simple choice is the negative classification loss, i.e.,
\begin{equation}
\label{eq:g1}
\gamma(\theta,\phi,x,y) = -\ell^{(\text{CE})}(g_\phi(f_\theta(x)), y).
\end{equation}
\noindent Inserting Eq.~\ref{eq:g1} into Eq.~\ref{eq:omega}, we have
\begin{equation}
\label{eq:omega1}
\begin{split}
\underset{\omega}{\operatorname{min}}~~~  \sum_{D_j\in \mathcal{D}_\text{val}}\sum_{d_j\in D_j}\tanh(&\ell^{(\text{CE})}(g_{\phi_j}(f_\theta^{(\text{NEW})}(x^{(j)})), y^{(j)})\\ -& \ell^{(\text{CE})}(g_{\phi_j}(f_\theta^{(\text{OLD})}(x^{(j)})), y^{(j)}))    
\end{split}
\end{equation}
We can now introduce the meta-loss $\ell^{(\text{Meta})}$ on $\mathcal{D}_\text{val}$ to abbreviate Eq.~\ref{eq:omega1} as $\underset{\omega}{\operatorname{min}}~ \ell^{(\text{Meta})}$.
Note that the design of $\gamma$ should reflect the demands of the testing stage. Here we choose to use classification loss because we assume the model will be deployed for a classification task eventually. An alternative choice could be a metric-based loss if we knew the final task was about retrieval. We emphasise that it is not necessary for the objective function used for the training sets (e.g., $\ell^{(\text{CE})}$) to match with the $\gamma$ function.

\subsection{Designing Feature-Critic $h_\omega$}
\label{sec:h}
Finally, we design our feature-critic network $h_\omega$ (i.e., $\ell^{(\text{Aux})}$). Recall the requirements for such an auxiliary loss: (i) It outputs a non-negative scalar; (ii) Its input depends on $\theta$. We note that MetaReg~\cite{Balaji18}, has a regularisation function that plays the similar role to $h_\omega$. MetaReg proposed the following form: $h_\omega(\theta) = \sum_{i} \omega_i |\theta_i|$. However, this introduces the same number of parameters as $\theta$. Doubling the number of model parameters in large modern CNNs is an expensive proposition that increases optimisation difficulty and overfitting risk. 

Therefore rather than designing $h_\omega$ to take $\theta$ directly, we propose a more efficient and effective way to enable $h_\omega$ to promote the base network's generalisation. Specifically, the auxiliary loss operates on the extracted features $f_\theta(x)$. Since our auxiliary generalisation-promoting loss operates on the feature representation produced by the base network, we denote it \modelname. 

Denote $F=f_\theta(X^{(j)})$ as the  $M\times H$ sized matrix stacking the $H$-dimensional features from $M$ examples in a mini-batch from the $j$th domain in the virtual training set  $\mathcal{D}_{\text{trn}}$. $h_\omega:=h_\omega(F)=h_\omega(f_\theta(X^{(j)}))$. A key requirement of $h_\omega$ is that it should be permutation invariant to the rows of $F$, i.e., it should not make a difference if we feed images indexed $[1,2,3]$ or $[3,2,1]$. Two available choices are:

(i) The set embedding \cite{NIPS2017_6931}, i.e.,
\begin{equation}
\label{eq:h1}
h_\omega(F)=\frac{1}{M}\sum_{i=1}^{M} \operatorname{MLP}_\omega(F_i)
\end{equation}
\noindent where $F_i$ denotes a row of $F$, and $\operatorname{MLP}$ is a multi-layer perceptron.

(ii) The flattened covariance matrix, i.e.,
\begin{equation}
\label{eq:h2}
h_\omega(F) = \operatorname{MLP}_\omega(\operatorname{Flatten}(F^T F))
\end{equation}
Finally, the MLP's output should be a scalar and we place a softplus activation to make sure its output is non-negative.

\subsection{Summary}

Bringing all the components together, we have the full Algo.~\ref{alg2}. To summarise, we randomly draw train/validation domains in each iteration and: Perform a putative feature extractor update on $\theta$ with and without the  auxiliary \modelname{} loss. Then generate a meta-loss based on whether or not the feature extractor update has improved performance on the validation set. Finally the feature extractor/classifier are updated using the supervised and auxiliary losses, and auxiliary loss itself is updated using the meta-loss.
\setlength{\textfloatsep}{2.5em}
\begin{algorithm}[t]
	\DontPrintSemicolon
	\KwIn{$\mathcal{D} = \{D_1, D_2, \dots, D_N\}$, $\alpha$, and $\eta$}
	\KwOut{$\theta$}
	\Begin{
		\While{not converge or reach max steps}{
			\textbf{Randomly split} $\mathcal{D}$:
			
			$\mathcal{D}_\text{trn} \cap \mathcal{D}_\text{val} = \emptyset$
			
			$\mathcal{D}_\text{trn} \cup \mathcal{D}_\text{val} = \mathcal{D}$
			
			\For{$t\in [1,2,\dots,T]$}{
				\textbf{Meta-train:}

				Sample mini-batch $d_{\text{trn}}$\cut{$=\{x^{(j)}, y^{(j)}\}$} from each $D_j\in \mathcal{D}_\text{trn}$
				
				$\ell^{(\text{CE})}(d_{\text{trn}}) \leftarrow$  Eq.~(\ref{eq:basic}) \hfill//Supervised  loss

				$\ell^{(\text{Aux})}(d_{\text{trn}}) \leftarrow$ Eqs.~(\ref{eq:h1} or \ref{eq:h2}). \hfill//Auxiliary loss
				
				$\theta^{(\text{OLD})} = \theta -  \alpha\nabla_\theta \ell^{(\text{CE})}$

				$\theta^{(\text{NEW})} = \theta^{(\text{OLD})} - \alpha\nabla_\theta \ell^{(\text{Aux})}$
				
				\textbf{Meta-test:}

				Sample mini-batch $d_{\text{val}}$\cut{$=\{x^{(k)}, y^{(k)}\}$} from each $D_k\in \mathcal{D}_\text{val}$
				
				$\ell^{\text{(Meta)}}(d_{\text{val}},\theta^{\text{(OLD)}},\theta^{\text{(NEW)}})\leftarrow$ Eq.~(\ref{eq:omega1})
				
				\hfill//Meta-loss

				\textbf{Meta-optimization:}
				
				$\theta \leftarrow \theta - \eta (\nabla_\theta \ell^{(\text{CE})} + \nabla_\theta \ell^{(\text{Aux})})$ \hfill//Update feat.
				
				$\phi \leftarrow \phi - \eta \nabla_\phi \ell^{(\text{CE})}$ \hfill//Update classifier
				
				$\omega \leftarrow \omega - \eta \nabla_\omega \ell^{\text{(Meta)}}$ \hfill//Update feature-critic
				
			}
		}
	}
	\caption{Full Algorithm}\label{alg2}
\end{algorithm}

\section{Experiments}
We evaluate our approach, first on the heterogeneous DG problem using the VD benchmark (Section~\ref{sec:expHetero}), and then on the conventional homogeneous DG using Rotated MNIST and PACS (Section~\ref{sec:expHomo}). Our demo code can be viewed on \url{https://github.com/liyiying/Feature\_Critic}.

\subsection{Heterogeneous DG experiments with VD}\label{sec:expHetero}
\keypoint{Dataset} The Visual Decathlon dataset, initially proposed for multi-domain learning \cite{Rebuffi17}, also provides a large scale and rigorous benchmark for DG. VD contains ten diverse domains including handwritten characters, pedestrians, traffic signs, etc. The images have been pre-processed to $72\times 72$. To use this benchmark for DG, we aim to train a network on a subset of source domains, and produce a robust feature extractor that provides a good representation for classification in a disjoint subset of target domains. It should do so `out-of-the-box', without further fine tuning. Specifically, we take the six larger datasets (CIFAR-100, Daimler Ped, GTSRB, Omniglot, SVHN and ImageNet) as source domains and hold out the four smaller datasets (Aircraft, D. Textures, VGG-Flowers and UCF101) as target domains. We use ImageNet pre-trained ResNet-18 \cite{He16} as the base network for all competitors. For computational efficiency, we freeze the first four blocks of ResNet-18 and only update the remaining blocks\nice{, as well as the average pooling layer,} during DG training. For all methods, the final feature is used to train SVM or KNN for the target task. All methods are evaluated by both average multi-class classification accuracy in the target domains, as well as the VD-Score metric \cite{Rebuffi17} that rewards consistently high performance across all domains.

\keypoint{Competitors} Few competitors can address heterogeneous DG. For these we consider AGG baseline (Eq~\ref{eq:basic}), CrossGrad \cite{Shankar18}, MetaReg \cite{Balaji18}, and Reptile \cite{nichol2018reptileFOML}. MetaReg is originally designed to produce a robust classifier given a fixed feature. We modify MetaReg to support the heterogeneous DG by (i) applying it on the feature extractor instead (as per our \modelname), called MR; (ii) applying it on the final layer of feature extractor, called MR-FL. Meanwhile Reptile is designed for few-shot meta-learning. However after modifying it for multi-domain rather than multi-task meta-learning, we found it effective for heterogeneous DG.

\keypoint{Feature-Critic Settings} We use the set embedding architecture for the critic network (Eq \ref{eq:h1}), as the covariance architecture requires too many parameters using high dimensional ResNet. During each iteration, we randomly choose four of the six source domains as meta-train, and the remaining two provide the meta-test (validation) domains. We train all components end-to-end using the AMSGrad \cite{AMSGrad} (batch-size/per meta-train domain=64, batch-size/per meta-test domain=32, lr=0.0005, weight decay=0.0001) for 30k iterations where the lr decayed in 5K, 12K, 15K, 20K iterations by a factor 5, 10, 50, 100, respectively. Similar to MetaReg \cite{Balaji18}, after the parameters are trained via meta-learning, we fine-tune the network on all source datasets for the final 10k iterations.

\begin{table*}[t]
\centering
\caption{Recognition accuracy ($\%$) and VD scores on four held out target datasets in Visual Decathlon using ResNet-18 extractor.}\label{tab:mainVD}
\resizebox{\textwidth}{!}{%
\begin{tabular}{c|ccccccc|ccccccc}
\toprule
\multirow{2}{*}{Target} & \multicolumn{7}{c|}{SVM Classifier} & \multicolumn{7}{c}{KNN Classifier}\\
 & Im.N. PT & CrossGrad & MR & MR-FL & Reptile & AGG &  FC & Im.N. PT & CrossGrad & MR & MR-FL & Reptile & AGG & FC\\
\hline
Aircraft & 16.62 & 19.92 & 20.91 & 18.18 & 19.62 & 19.56 & \textbf{20.94} & 11.46 & 15.93 & 12.03 & 11.46 & 13.27 & 14.03 & \textbf{16.01}\\
D. Textures & \textbf{41.70} & 36.54 & 32.34 & 35.69 & 37.39 & 36.49 & 38.88 & 39.52 & 31.98 & 27.93 & \textbf{39.41} & 32.80 & 32.02 & 34.92\\
VGG-Flowers & 51.57 & 57.84 & 35.49 & 53.04 & 58.26 & 58.04 & \textbf{58.53} & 41.08 & \textbf{48.00} & 23.63 & 39.51 & 45.80 & 45.98 & 47.04\\
UCF101 & 44.93 & 45.80 & 47.34 & 48.10 & 49.85 & 46.98 & \textbf{50.82} & 35.25 & 37.95 & 34.43 & 35.25 & 39.06 & 38.04 & \textbf{41.87}\\
\hline
Ave. & 38.71 & 40.03 & 34.02 & 38.75 & 41.28 & 40.27 & \textbf{42.29} & 31.83 & 33.47 & 24.51 & 31.41 & 32.73 & 32.52 & \textbf{34.96}\\
\hline
VD-Score & 308 & 280 & 269 & 296 & 324 & 290 & \textbf{344} & 215 & 188 & 144 & 215 & 201 & 189 & \textbf{236}\\
\bottomrule
\end{tabular}%
}
\end{table*}

\keypoint{Results}
We first assume the full training split is available for each target domain. Table~\ref{tab:mainVD} shows that: (i) The original ImageNet feature transfers to novel tasks reasonably well, as observed by classic studies \cite{yosinski2014howTransferable}. (ii) Demonstrating the benefit of simply exploiting large datasets, the  AGG baseline's feature, trained on more than 1.39 million images across the six domains, provides strong performance. However, while it has a higher average accuracy than the ImageNet feature, AGG's VD score is lower, reflecting its inconsistent performance. Thus obtaining consistently high scores from multi-domain training is non-trivial. Naively aggregating more diverse source domains into training can both help and hinder performance (for example, depending on if aggregated domains are particularly similar or dissimilar to a given target). Nevertheless, AGG sometimes outperforms prior purpose designed DG methods CrossGrad and MetaReg, with only Reptile producing a feature that outperforms AGG in both accuracy and VD-score metrics. (iii) Overall, our  \modelname{} (FC) method generally provides the best performance across domains and across both types of classifiers evaluated. 

Although the above application scenario of heterogeneous DG is one where compute, memory or human resources rule out feature fine-tuning, another motivating scenario is where the target domain data is too sparse for effective fine-tuning. Thus we next investigate the situation if less target data is available. Specifically, we repeat the evaluation assuming that  [10$\%$, 25$\%$, 50$\%$, 100$\%$] of the training split is available for SVM/KNN training. Table \ref{tab:difpro} reports target domain test accuracies under these settings. We can see that \modelname{} provides a consistent improvement over the alternatives. Finally, we also consider a genuinely few-shot setting for the target domain. In this case we consider $K=3,5,8,10$ labelled examples per class in the target domain, and perform KNN recognition on their test sets. The results in Table~\ref{tab:K-shot} show that for simple similarity-based matching in a novel target domain, \modelname{} also provides the best off-the-shelf feature representation.

In summary the \modelname{} meta-training strategy produces a feature extractor that is generally useful for diverse target problems in an off-the-shelf feature + shallow classifier configuration. The results outperform both the standard ImageNet feature and the obvious Data Aggregation extension across a range of operating points in the target domain from the few to many-shot regime. This suggests that \modelname{} trained feature extractors are of wide potential value in diverse applications.

\begin{table*}[htb]
\caption{VD recognition accuracy differences ($\%$) against AGG with different proportions of training data available.}\label{tab:difpro}
\resizebox{\textwidth}{!}{
\centering
\begin{tabular}{c|ccccccc|ccccccc}
\toprule
\multirow{2}{*}{Data} & \multicolumn{7}{c|}{SVM Classifier} & \multicolumn{7}{c}{KNN Classifier}\\
 & AGG & Im.N. PT & CrossGrad & MR & MR-FL & Reptile & FC & AGG & Im.N. PT & CrossGrad & MR & MR-FL & Reptile & FC\\
\hline
10$\%$ & 18.27 & +1.74 & +0.58 & -2.38 & -1.50 & \textbf{+1.83} & +1.31 & 13.93 & +1.15 & -0.29 & -3.08 & +0.63 & +1.01 & \textbf{+1.39}\\
25$\%$ & 30.10 & -5.14 & -2.48 & -8.44 & -2.48 & -2.13 & \textbf{+0.87} & 23.80 & -3.62 & -0.06 & -9.24 & -3.78 & -1.32 & \textbf{+1.48}\\
50$\%$ & 34.63 & -3.55 & -0.89 & -7.04 & -0.07 & +0.37 & \textbf{+2.12} & 30.19 & -5.02 & -0.41 & -8.87 & -4.17 & -0.15 & \textbf{+2.60}\\
100$\%$& 40.27 & -1.56 & -0.24 & -6.25 & -1.52 & +1.01 & \textbf{+2.02} & 32.52 & -0.69 & +0.95 & -8.01 & -1.11 & +0.21 & \textbf{+2.44}\\
\hline
Ave.   & 30.82 & -2.13 & -0.76 & -6.03 & -1.39 & +0.27 & \textbf{+1.58} & 25.11 & -2.05 & +0.05 & -7.30 & -2.11 & -0.06 & \textbf{+1.98}\\
\bottomrule
\cut{
\toprule
Data & AGG & Im.N. PT & CrossGrad & MetaReg & Reptile & \modelname{}\\
\hline
10$\%$ & 13.93 & +1.15 & -0.29 & -3.08 & +1.01 & \textbf{+1.39}\\
25$\%$ & 23.80 & -3.62 & -0.06 & -9.24 & -1.32 & \textbf{+1.48}\\
50$\%$ & 30.19 & -5.02 & -0.41 & -8.87 & -0.15 & \textbf{+2.60}\\
100$\%$& 32.52 & -0.69 & +0.95 & -8.01 & +0.21 & \textbf{+2.44}\\
\hline
Ave.   & 25.11 & -2.05 & +0.05 & -7.30 & -0.06 & \textbf{+1.98}\\
\bottomrule
}
\end{tabular}}
\end{table*}

\begin{table*}[t]
\centering
\caption{Recognition accuracy ($\%$) averaged over 10 test runs on VD K-shot learning.}
\resizebox{0.85\textwidth}{!}{
\begin{tabular}{c|ccccccc}
\toprule
K & Im.N. PT & CrossGrad & MR & MR-FL & Reptile & AGG & \modelname\\
\hline
3 & 13.88 $\pm$ 1.82 & 14.01 $\pm$ 1.98 & 8.86 $\pm$ 1.52 & 13.26 $\pm$ 1.62 & 14.59 $\pm$ 2.30 & 13.75 $\pm$ 1.77 & \textbf{15.18} $\pm$ 2.26\\
5 & 17.63 $\pm$ 1.55 & 17.74 $\pm$ 1.40 & 12.01 $\pm$ 1.51 & 17.22 $\pm$ 0.84 & \textbf{19.17} $\pm$ 0.94 & 18.00 $\pm$ 1.58 & 19.02 $\pm$ 1.57\\
8 & 21.58 $\pm$ 1.12 & 20.98 $\pm$ 1.07 & 14.37 $\pm$ 1.35 & 21.61 $\pm$ 0.92 & 21.24 $\pm$ 1.40 & 21.36 $\pm$ 1.05 & \textbf{22.39} $\pm$ 1.00\\
10 & 23.40 $\pm$ 0.99 & 22.84 $\pm$ 0.89 & 15.61 $\pm$ 0.88 & 22.80 $\pm$ 0.61 & 23.42 $\pm$ 1.12 & 22.56 $\pm$ 0.84 & \textbf{24.23} $\pm$ 1.00\\
\hline
Ave. & 19.12 & 18.89 & 12.71 & 18.72 & 19.61 & 18.92 & \textbf{20.21}\\
\bottomrule
\end{tabular}}
\label{tab:K-shot}
\end{table*}

\subsection{Homogeneous DG experiments}\label{sec:expHomo}

\subsubsection{Rotated MNIST}
\keypoint{Dataset and Settings} Rotated MNIST \cite{Ghifary15} contains six domains with each corresponding to a degree of roll rotation in the classic MNIST dataset. The basic view (M0) is formed by randomly choosing 100 images each of ten classes from the original MNIST and we create 5 rotating domains from M0 with $15^\circ$ rotation each in clockwise direction, denoted by M15, M30, M45, M60, and M75. Following the setting in \cite{Shankar18}, we perform leave-one-domain-out experiments by picking one domain to hold out as the target. We compare AGG baseline, as well as CrossGrad and MetaReg. For a recognition network all competitors use the standard MNIST CNN with two conv and one FC layer as the feature network and another FC layer as the classifier. We note prior studies \cite{Ghifary15,Shankar18,Deshmukh17} did not release specific selection of digits from within MNIST, so our results do not match the numbers in those papers exactly. However, we repeat all experiments 10 times and report the mean and standard deviation of recognition accuracy. 

For \modelname{}, we train using the AMSGrad optimizer (lr=0.001, weight decay=0.00005) for 5,000 iterations. For each iteration, one meta-train and one meta-test domain are chosen randomly from the five source domains. We also use this opportunity to compare the two variants of our loss function: \modelname{}-MLP and \modelname{}-Flatten.

\keypoint{Results} 
It can be seen from Table \ref{rMNIST} that AGG is again a strong baseline to beat. Over ten trials of 1000 digit samples, CrossGrad and MetaReg failed to match AGG, with only Reptile matching AGG's performance. Meanwhile, \modelname{} performs well with both variants of the auxiliary loss network, with the set embedding (Eq.~\ref{eq:h1}) performing slightly better than the covariance matrix embedding (Eq.~\ref{eq:h2}).

\begin{figure}[tb]
\centering
\includegraphics[width=0.24\textwidth]{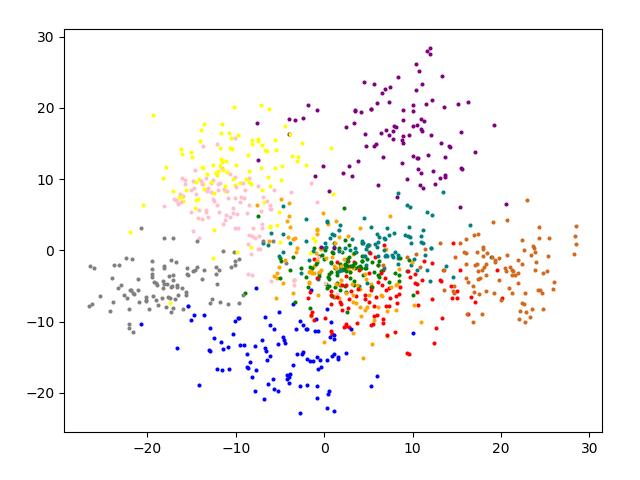}~
\includegraphics[width=0.24\textwidth]{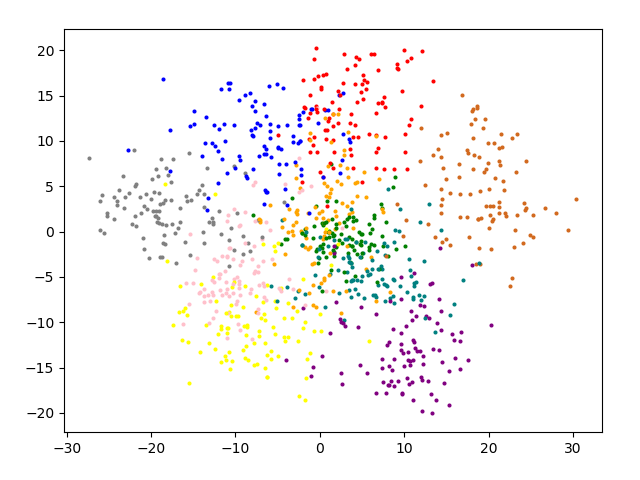}
\caption{ Rotated MNIST. PCA projections of target domain M15 features. Left: AGG. Right: \modelname{}. Color: Digit.} \label{fig:3}
\end{figure}

To qualitatively visualise the results we perform PCA projections of the features in the target domain. Figure~\ref{fig:3} shows these projections, taking as an example the M15 domain as held out. Each dot denotes an image and the colour denotes its label. We can see that \modelname{} (take MLP style as an example) feature extractor provides improved separability in the target domain compared to the AGG baseline. 

\begin{figure}[t]
\centering
\includegraphics[width=0.15\textwidth]{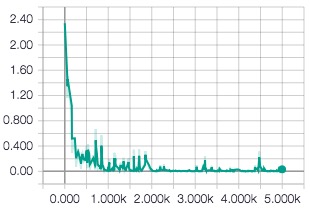}~
\includegraphics[width=0.15\textwidth]{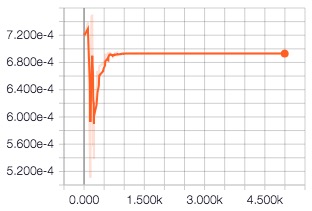}~
\includegraphics[width=0.15\textwidth]{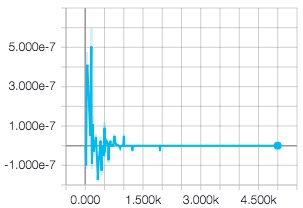}
\includegraphics[width=0.155\textwidth]{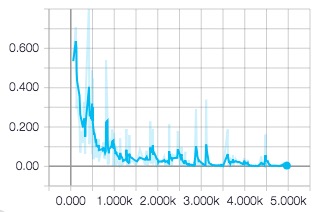}~
\includegraphics[width=0.152\textwidth]{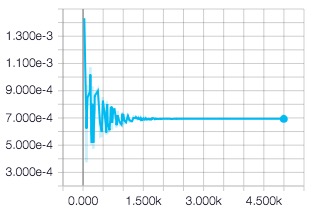}~
\includegraphics[width=0.1513\textwidth]{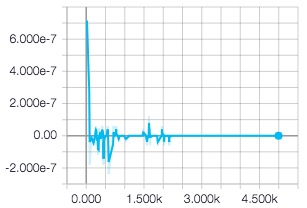}
\caption{Loss curves of \modelname{} on Rotated MNIST. Left to right: CE loss, auxiliary loss and meta-loss during DG training. Top: \modelname-MLP. Bottom: \modelname-Flatten.} \label{fig:2}
\end{figure}

\keypoint{Further Analysis} 
Figure \ref{fig:2} reports the loss curves of cross entropy loss, auxiliary loss, and meta-loss for \modelname{} during training. The cross entropy loss converges to zero, as the network usually can fit the training data perfectly. The auxiliary loss fluctuates up and down for the early stage of training, and finally stabilises to a small value. Because the auxiliary loss function $h_\omega$ itself is learned, its behaviour changes with its own learning process, which explains the fluctuations, esp. for the early stage. It is more interesting to see the pattern of meta-loss, which is the performance difference of feature extractor parameterised by $\theta^{(\text{OLD})}$ and that by $\theta^{(\text{NEW})}$. If we select zero as a threshold, meta-loss has a clear pattern: ``above zero'' $\rightarrow$ ``below zero'' $\rightarrow$ ``'being zero'. This pattern is expected because: (i) For the early stage, the auxiliary loss' parameters are randomly initialised, so it knows little about how to help generalise, thus the gradients produced by it are rather random and less likely to help. Thus $\theta^{(\text{OLD})}$-based model outperforms $\theta^{(\text{NEW})}$-based model. (ii) With the updating of $\omega$, $h_\omega$ improves and begins to make $\theta^{(\text{NEW})}$ better than $\theta^{(\text{OLD})}$. During this period, gradients produced by the auxiliary loss help the model learn to generalise. (iii) For the late stage, meta-loss goes towards zero, which indicates that $h_\omega$ no longer helps (but it does not hurt either), as all of its knowledge has now been distilled into the feature extractor. The pattern of the three losses also demonstrates that, empirically, the whole algorithm converges, including the learned auxiliary loss.

\begin{table*}[t]
\centering
\caption{Recognition accuracy ($\%$) averaged over 10 train+test runs on Rotated MNIST.}
\resizebox{0.9\textwidth}{!}{
\begin{tabular}{c|cccccc}
\toprule
Target & CrossGrad & MetaReg & Reptile & AGG & \modelname-MLP & \modelname-Flatten \\
\hline
M0 & 86.03 $\pm$ 0.69 & 85.70 $\pm$ 0.31 & 87.78 $\pm$ 0.30 & 86.42 $\pm$ 0.24 & \textbf{89.23} $\pm$ 0.25 & 87.04 $\pm$ 0.31\\
M15 & 98.92 $\pm$ 0.53 & 98.87 $\pm$ 0.41 & 99.44 $\pm$ 0.22 & 98.61 $\pm$ 0.27 & \textbf{99.68} $\pm$ 0.24 & 99.53 $\pm$ 0.27\\
M30 & 98.60 $\pm$ 0.51 & 98.32 $\pm$ 0.44 & 98.42 $\pm$ 0.24 & 99.19 $\pm$ 0.19 & 99.20 $\pm$ 0.20 & \textbf{99.41} $\pm$ 0.18\\
M45 & 98.39 $\pm$ 0.29 & 98.58 $\pm$ 0.28 & 98.80 $\pm$ 0.20  & 98.22 $\pm$ 0.24 & 99.24 $\pm$ 0.18 & \textbf{99.52} $\pm$ 0.24\\
M60 & 98.68 $\pm$ 0.28 & 98.93 $\pm$ 0.32 & 99.03 $\pm$ 0.28 & 99.48 $\pm$ 0.19 & \textbf{99.53} $\pm$ 0.23 & 99.23 $\pm$ 0.16\\
M75 & 88.94 $\pm$ 0.47 & 89.44 $\pm$ 0.37 & 87.42 $\pm$ 0.33 & 88.92 $\pm$ 0.43 & 91.44 $\pm$ 0.34 & \textbf{91.52} $\pm$ 0.26\\
\hline
Ave. & 94.93 & 94.97 & 95.15 & 95.14 & \textbf{96.39} & 96.04\\
\bottomrule
\end{tabular}}
\label{rMNIST}
\end{table*}

\begin{table*}[tb]
\centering
\footnotesize
\caption{Cross-domain recognition accuracy ($\%$) on PACS using train split  \cite{Li17} for training.}\label{tab:PACS}
\resizebox{0.9\textwidth}{!}{
\begin{tabular}{p{0.7cm}<{\centering}|p{0.6cm}<{\centering}p{1.3cm}<{\centering}p{0.6cm}<{\centering}p{1.2cm}<{\centering}p{0.7cm}<{\centering}p{0.7cm}<{\centering}p{0.8cm}<{\centering}p{0.7cm}<{\centering}p{1.0cm}<{\centering}p{0.8cm}<{\centering}p{0.6cm}<{\centering}p{0.6cm}<{\centering}}
\toprule
 Target & DICA& \cut{LRE-SVM &} D-MTAE & DSN & TF-CNN & MLDG & DANN & CIDDG & Reptile & CrossGrad & MetaReg & AGG & FC\\
\hline
 A & 64.6 \cut{& 59.7} & 60.3 & 61.1 & 62.9 & \textbf{66.2} & 63.2 & 62.7 & 63.4 & 61.0 & 63.5 & 63.3 & 64.4\\
 C & 64.5 \cut{& 52.8} & 58.7 & 66.5 & 67.0 & 66.9 & 67.5 & \textbf{69.7} & 67.5 & 67.2 & 69.5 & 66.3 & 68.6\\
 P & \textbf{91.8} \cut{& 85.5} & 91.1 & 83.3 & 89.5 & 88.0 & 88.1 & 78.7 & 88.7 & 87.6 & 87.4 & 88.6 & 90.1\\
 S & 51.1 \cut{& 37.9} & 47.9 & 58.6 & 57.5 & 59.0 & 57.0 & \textbf{64.5} & 55.9 & 55.9 & 59.1 & 56.5 & 58.4\\
\hline
Ave. & 68.0 \cut{& 59.0} & 64.5 & 67.4 & 69.2 & 70.0 & 69.0 & 68.9 & 68.9 & 67.9 & 69.9 & 68.7 & \textbf{70.4}\\
\bottomrule\end{tabular}}
\label{PACS_train}
\end{table*}

\begin{table}[tb]
\centering
\caption{Cross-domain recognition accuracy ($\%$) on PACS using train+validation split \cite{Balaji18} for training.}\label{tab:PACS2}
\resizebox{1.0\columnwidth}{!}{
\begin{tabular}{c|ccccc}
\toprule
Target & CrossGrad & MetaReg & Reptile & AGG & \modelname\\
\hline
A & 64.84 & \textbf{69.82} & 64.35 & 63.77\cut{(67.21)} & 64.89 \\
C & 67.69 & 70.35 & 70.09 & 66.77\cut{(66.12)} & \textbf{71.72} \\
P & 88.48 & \textbf{91.07} & 88.78 & 88.62\cut{(88.47)} & 89.94 \\
S & 57.52 & 59.26 & 59.91 & 57.27\cut{(55.32)} & \textbf{61.85} \\
\hline
Ave. & 69.63 & \textbf{72.62} & 70.78 & 69.11\cut{(69.28)} & 72.10 \\
\bottomrule
\end{tabular}}
\end{table}

\subsubsection{Evaluation on PACS dataset}
\keypoint{Dataset and Settings} PACS \cite{Lida17} is a recent object recognition benchmark for domain generalisation. PACS contains 9991 images of size $224\times 224$ from four different domains - Photo, Art painting, Cartoon and Sketch. It has 7 categories across these domains: dog, elephant, giraffe, guitar, house, horse and person. We follow the standard protocol and perform leave-one-domain-out evaluation. Beyond this there have been two splits of PACS used in the literature.  PACS was defined with a train/validation/test split within each domain. In \citet{Li18} models are trained on the train split alone with the validation split used for early stopping. In \citet{Balaji18} the combined train+validation splits were used to train the models, resulting in slightly higher performance due to more data. For direct comparison with previously published results we evaluate both of these settings.

The ImageNet pre-trained AlexNet \cite{Krizhevsky12} is used as the backbone network. Our competitors include: DICA \cite{Muandet13}, D-MTAE \cite{Ghifary15}, DSN \cite{Bousmalis16}, TF-CNN \cite{Lida17}, MLDG \cite{Li18}, DANN \cite{Ganin16}, CIDDG \cite{LiY18conditional}, Reptile \cite{nichol2018reptileFOML}, CrossGrad \cite{Shankar18} and MetaReg \cite{Balaji18}. We note that DANN is designed for domain adaptation, and Reptile for few-shot learning. We re-purpose them for DG. DANN, Reptile,  CrossGrad, AGG, and MetaReg in Table~\ref{tab:PACS} are our implementations. The other results are taken from \citet{Li18}, \citet{LiY18conditional} and \citet{Balaji18}. Among these, MetaReg makes a domain general classifier; MLDG aligns gradients to achieve a more robust optima; CrossGrad synthesises data for a new domain; DANN makes indistinguishable representations across source domains; CIDDG learns discriminative features to match the distributions across domains; \modelname{} (FC) learns representations that generalise to new domains with a better objective since it trains a supervised loss and explicitly simulates domain shift.

Our \modelname{} (set embedding variant) is trained with M-SGD optimizer (batch size/per meta-trian domain=32, batch size/per meta-test domain=16, lr=0.0005, weight decay=0.00005, momentum=0.9) for 45K iterations. At each iteration, we randomly choose two of the three source domains as meta-train and the remaining one as meta-test.
 
\keypoint{Results} The comparison with state-of-the-art methods on PACS dataset is shown in Table~\ref{tab:PACS} and Table~\ref{tab:PACS2}. AGG provides a hard baseline to beat as usual. Nevertheless \modelname{} performs comparably to the best performing state of the art alternative in both settings of this benchmark.

\section{Conclusion}
We addressed the domain generalisation problem with a particular focus on the heterogeneous case,  by meta-learning a regulariser to help train a feature extractor to be domain invariant. The resulting feature extractor outperforms alternatives for general purpose use as a fixed downstream image encoding. Evaluated on Visual Decathlon -- the largest DG evaluation thus far -- this suggests that \modelname{} trained feature extractors could be of wide potential value in diverse applications. Furthermore \modelname{} also performs favourably compared to state-of-the-art in the homogeneous DG setting. In future work we will apply \modelname{} to other problems including RL, and explore the impact on fine-tuning target problems.

\section*{Acknowledgements}
This work was produced while the first author was
visiting the University of Edinburgh. This work was supported by National Natural Science Foundation of China (Grant No. 61751208),
 China Scholarship Council, EPSRC grant EP/R026173/1, and NVIDIA Corporation GPU donation.


\bibliography{example_paper}
\bibliographystyle{icml2019}

\end{document}